\documentclass{article}

\usepackage{microtype}
\usepackage{graphicx}
\usepackage{subfigure}
\usepackage{booktabs} 

\usepackage{hyperref}

\usepackage[accepted]{icml2021}

\usepackage{amsmath,amsfonts}
\usepackage{url}
\usepackage{comment}

\newcommand{\bmat}[1]{\begin{bmatrix}#1\end{bmatrix}} %

\renewcommand{\b}[1]{\ensuremath{\mathbf{#1}}} 

\newcommand{\biidx}[2]{\ensuremath{{\mathbf{#2}}_{#1}}}

\newcommand{\bkex}[1]{\ensuremath{{\mathbf{#1}}^{(k)}}} 
\newcommand{\bkoex}[1]{\ensuremath{{\mathbf{#1}}^{(k+1)}}} 

\newcommand{\tp}{^{\top}}   

\icmltitlerunning{Stochastic NLLS}
\begin{document}

\twocolumn[
\icmltitle{Nonlinear Least Squares for  
        Large-Scale Machine Learning \\
        using Stochastic Jacobian Estimates}



\icmlsetsymbol{equal}{*}

\begin{icmlauthorlist}
\icmlauthor{Johannes J.~Brust}{ucsd}
\end{icmlauthorlist}

\icmlaffiliation{ucsd}{Department of Mathematics, University of California San Diego, San Diego, CA}

\icmlcorrespondingauthor{Johannes J. Brust}{jjbrust@ucsd.edu}

\icmlkeywords{Machine Learning, ICML}

\vskip 0.3in
]



\printAffiliationsAndNotice{\icmlEqualContribution} 

\begin{abstract}
For large nonlinear least squares loss functions in machine learning
we exploit the property that the number of model parameters typically
exceeds the data in one batch. This implies a low-rank structure in the
Hessian of the loss, which enables effective means to compute search
directions. Using this property, we develop two algorithms that estimate Jacobian
matrices and perform well when compared to state-of-the-art methods.
\end{abstract}

\section{Introduction}
\label{sec:intro}

A major challenge in optimization of machine learning models is the 
large size of both; the large number of model parameters and the number of 
data points. Therefore, the use of higher derivative information whose 
complexity may scale quadratically with problem size typically has to be done in a careful manner \cite{BottouCurtisNocedal18,XuRoostaMahoney20}. Moreover, because very large data amounts may not be manageable all at once, even methods that use $1^{\textnormal{st}}$ order gradients only, usually randomly split the whole dataset into smaller
blocks (minibatches). Once 
the data is appropriately prepared, the goal of an effective optimization
algorithm is to fit the model accurately to the data by iteratively minimizing
a loss objective function. A commonly used loss measure is the
(mean) sum-of-squares \cite{HighamHigham19,SraNowozinWright11}. Since machine learning models are
normally nonlinear, the resulting optimization problems are formulated as
nonlinear least squares (NLLS) problems:
\begin{equation}
    \label{eq:nlls}
    \underset{ \b{w} \in \mathbb{R}^n }{ \textnormal{ minimize } }
    \frac{1}{\gamma} 
    \sum_{i=1}^N (m(\b{w};\biidx{i}{x}) - y_i )^2,
\end{equation}
where $ \gamma > 0 $ is a scaling value, $ m(\cdot,\b{x}) $ is the machine
learning model, $ (\biidx{i}{x}, y_i)_{i=1}^N $ are ``(feature, label)'' data pairs
and $ \b{w} $ are the optimization weights. Note that in \eqref{eq:nlls}
both $n$ and $N$ can be large. By splitting the data pairs into $B$ 
minibatch blocks each containing $L$ elements (i.e., $N=BL$) we define the
indices $ i = bL + l \equiv i(l,b) $ for $ 1 \le b \le B $ and $ 1 \le l \le L $.
Moreover, defining the residual at a data pair as
$ r_i \equiv (m(\b{w},\biidx{i}{x}) - y_i ) $ and the vector of 
residuals in a batch
$
    \biidx{b}{r} \equiv \bmat{ r_{i(1,b)},r_{i(2,b)} \cdots, r_{i(L,b)} }\tp,
$
the least squares objective is
\begin{equation}
    \label{eq:nlls1}
    \sum_{i=1}^N (m(\b{w};\biidx{i}{x}) - y_i )^2 =
    \sum_{i=1}^N r_i^2 =
    L \sum_{b=1}^B \frac{1}{L} \biidx{b}{r}\tp \biidx{b}{r}. 
\end{equation}
In the remainder we assume that for each batch the value of the
objective $ \biidx{b}{r}\tp \biidx{b}{r} \big / L $ and its corresponding gradient 
$ \biidx{b}{g} \equiv \nabla_w (\biidx{b}{r}\tp \biidx{b}{r}) \big / L $
are known.

\section{Contribution}
\label{sec:contribution}
This work exploits the property that usually the number of data pairs
$L$ within a minibatch is smaller than the number of model parameters,
i.e., $L < n$. Combining this with the nonlinear least squares 
objective \eqref{eq:nlls1}, we note that $2^{\textnormal{nd}}$ derivative matrices contain
a low-rank structure that can be used for effective search direction computations. In particular, we describe two algorithms that
approximate higher derivative information using a low memory footprint,
and that have computational complexities, which are comparable to
state-of-the-art first order methods. Other work on stochastic NLLS problems
includes the computation of higher derivatives via a domain-decomposition process
\cite{HuangHuangSun21} or the analysis of noisy loss and residual values 
\cite{BergouDiouaneKungurtsevRoyer20}. An initial backpropagation algorithm for the
NLLS objective was developed in \cite{HaganMenhaj94}. However, none of these works
exploit the low rank structures in the loss objective. 

\section{Methods}
\label{sec:methods}
Suppose that at a given batch $1 \le b \le B$ the loss 
$ \biidx{b}{r}\tp \biidx{b}{r} \big / L $ and its gradient $ \biidx{b}{g} $
are computed. We temporarily suppress the subscripts so that the
residual vector and gradient are denoted $\b{r}$ and $\b{g}$, respectively.
The $1^{\textnormal{st}}$ and $2^{\textnormal{nd}}$ derivatives of the
loss are
\begin{align}
    \label{eq:derivsJ}
    \nabla \left(\frac{ \b{r}\tp \b{r} }{L} \right) &= 
    \frac{2}{L} (\nabla \b{r}) \b{r} = \b{g}, \quad \text{ and } \\ 
    \label{eq:derivsH}
    \nabla^2 \left(\frac{ \b{r}\tp \b{r} }{L}\right) &= 
    \frac{2}{L}\big( \nabla \b{r} (\nabla \b{r})\tp + \sum_{l=1}^L (\nabla^2 \biidx{l}{r}) \b{r}_l \big).
\end{align}
Throughout we will denote the Jacobian matrix by 
$ \nabla \b{r} \equiv \b{J} \in \mathbb{R}^{n \times L} $ and the
Hessian matrix as $ \nabla^2 (\b{r}\tp \b{r} / L) \equiv \b{H} $. 
We define a new search direction at a current iteration $k$ as the update to the weights: $ \bkoex{w} = \bkex{w} + \b{s} $. Using Newton's method the search direction is defined by the linear system
\begin{equation}
    \label{eq:newton}
    \bkex{H} \biidx{N}{s} = - \bkex{g},
\end{equation}
where gradient, Jacobian and Hessian are evaluated at the current iteration, i.e., $ \bkex{g} = \b{g}(\bkex{w}) $, $ \bkex{J} = \b{J}(\bkex{w}) $, and $ \bkex{H} = \b{H}(\bkex{w}) $. For large $n$ forming the Hessian matrix and
solving with it is, however, not feasible. Thus we estimate the components in
\eqref{eq:derivsH}. First, note that the residuals near the solution
are expected to be small and thus estimating the matrix
$\sum_{l=1}^L (\nabla^2 \biidx{l}{r}) \b{r}_l \approx \frac{1}{\alpha} \b{D}$ with a diagonal $ \b{D} $ and scalar $\alpha > 0$ appears reasonable. For regular least squares problems setting 
$ \b{D} = \b{0} $ yields a Gauss-Newton type method, whereas $ \b{D} \neq \b{0} $ corresponds to a Levenberg-Marquardt type method (cf. \cite{NocW06}). In this work we develop methods that estimate the components of the Hessian, which are suited for large data
in machine learning problems. In particular, we define search directions
by the system 
\begin{equation}
    \label{eq:search}
    \bigg( \bkex{J} {\bkex{J}}\tp + \frac{1}{\alpha} \bkex{D}  \bigg) \b{s} = - \bkex{g},
\end{equation}
where the Jacobian $ \bkex{J} $ is approximated. Since
$L < n $ the matrix in parenthesis in \eqref{eq:search} can be inverted using the 
Sherman-Morrison-Woodbury (SMW) inverse. Even so, computing the full
Jacobian $\bkex{J}$ is typically very expensive for large $L$ and $n$.
Therefore, we describe two approaches for approximating $ \bkex{J} $
that enable effective computation of $\b{s}$ in \eqref{eq:search}.
Specifically, we derive representations of $ \bkex{J} $
based on the gradient/Jacobian relation in \eqref{eq:derivsJ}.
In particular, the Jacobian matrices in this work approximate the relation
$ \frac{2}{L} \bkex{J} \bkex{r} = \bkex{g} $.

\subsection{Rank-1 Jacobian Estimate}
\label{sec:rank1}
Let $1 \le s \le k$ denote an intermediate iteration. In our 
first approach we define the Jacobian estimate by a rank-1 matrix. 
More specifically, we represent an intermediate Jacobian in the form
\begin{equation*}
\widehat{\b{J}}^{(s)}_1 =
\theta
    \bmat{
     \beta_1 \rho_1&\beta_1 \rho_2 & \cdots & \beta_1 \rho_L \\
     \beta_2 \rho_1 & \beta_2 \rho_2 & \cdots & \beta_2 \rho_L \\
     \vdots  & \vdots & \vdots & \vdots \\
     \beta_L \rho_1 & \beta_L \rho_2 & \cdots & \beta_L \rho_L \\
     \vdots  & \vdots & \vdots & \vdots \\
     \vdots  & \vdots & \vdots & \vdots \\
     \beta_{n} \rho_1 & \beta_{n} \rho_2 & \cdots & \beta_{n} \rho_L
    },
\end{equation*}
where the $n+L+1$ coefficients $ \beta_i, \rho_l, \theta $ are to be 
determined. In order to satisfy the gradient/Jacobian relation
$ \frac{2}{L} \widehat{\b{J}}^{(s)}_1 \b{r}^{(s)} = \b{g}^{(s)}  $
we deduce the form
\begin{equation}
    \label{eq:rank1a}
    \widehat{\b{J}}^{(s)}_1 = \frac{L}{2 \| \b{r}^{(s)} \|^2 }
                \b{g}^{(s)}{\b{r}^{(s)}}\tp.
\end{equation}
Thus the coefficients at an intermediate iterate are specified as $ \beta_i = \b{g}_i $, $ \rho_l = \b{r}_l $ and $ \theta = \frac{L}{2 \| \b{r} \|^2} $. Since the entire objective function in \eqref{eq:nlls} consists of
all data pairs beyond those in one batch, our rank-1 Jacobian approximation
accumulates previously computed information 
\begin{equation}
    \label{eq:rank1}
        \b{J}^{(k)}_1 = \left(\frac{1}{2f^{(k)}}\sum_{s=1}^k \b{g}^{(s)} \right)
        \bmat{{\b{r}^{(1)}}\tp &
                \hdots &
                {\b{r}^{(k)}}\tp},
\end{equation}
where
\begin{equation*}
    f^{(k)} = \sum_{s=1}^k \frac{1}{L} \| \b{r}^{(s)} \|^2.
\end{equation*}
Note that in the rank-1 approximation of \eqref{eq:rank1} the vectors
of residuals $ \bmat{{\b{r}^{(1)}}\tp &
                \hdots &
                {\b{r}^{(k)}}\tp} $ need not be explicitly stacked, because
the search direction in \eqref{eq:search} uses the term 
$ \bkex{J}_1 {\bkex{J}_1}\tp  $ in which the residuals reduce to
the inner product 
$ \sum_{s=1}^k {\b{r}^{(s)}}\tp {\b{r}^{(s)}} = L f_k $.

\subsection{Rank-L Jacobian Estimate}
\label{sec:rankL}
For our second approach we use the rank-L representation
\begin{equation}
    \label{eq:rankLa1}
\widehat{\b{J}}^{(s)}_L = \text{diag}(\b{g}^{(s)})
    {\b{P}_1}\tp
    \bmat{
     \beta_1    & & & \\
                & \beta_2   &  & \\
                &           & \ddots & \\
                & &         & \beta_L\\
                & &         & \beta_{L+1}\\
                & &         & \vdots\\
                & &         & \beta_n\\
    }\b{P}_2,
\end{equation}
where $ {\b{P}_1} $ and $ {\b{P}_2} $ are two random permutation matrices
that rearrange the row and column ordering. In order
to satisfy the relation $ \frac{2}{L} \widehat{\b{J}}^{(s)}_1 \b{r}^{(s)} = \b{g}^{(s)}  $, the Jacobian estimate $\widehat{\b{J}}^{(s)}_L$ can
be written as
\begin{equation}
    \label{eq:rankLa}
    \widehat{\b{J}}^{(s)}_L = \frac{L}{2} \text{diag}(\b{g}^{(s)}) {\b{P}_1} \tp \b{R}^{(s)} {\b{P}_2} 
\end{equation}
where $ \b{R}^{(s)} $ is the rectangular matrix in \eqref{eq:rankLa1} with
elements
\begin{equation*}
    \beta_i = \frac{1}{\b{r}^{(s)}_i}, \: 1\le i < L \quad \text{ and } \quad
    \beta_j = \frac{1}{\b{r}^{(s)}_L}, \: L\le j \le n.
\end{equation*}

We accumulate the intermediate estimates as to define the rank-L
Jacobian as
\begin{equation}
    \label{eq:rankL}
    \bkex{J}_L = \sum_{s=1}^k \widehat{\b{J}}^{(s)}_L.
\end{equation}

Finally, note that we do not form the Jacobians in
\eqref{eq:rank1} or \eqref{eq:rankL} explicitly. 
Rather, we accumulate the information in $ \b{g}^{(s)}, \| \b{r}^{(s)} \|^2 $
and $ \frac{\b{g}^{(s)}_i}{ \b{r}^{(s)}_l } $. For instance, we represent
$ \b{J}_1^{(k)}   $ by storing the vector $ \sum_{s=1}^k \b{g}^{(s)} $ and 
the scalar $f^{(k)}$. For $ \b{J}_L^{(k)} $ we store the non-zeros elements of 
\eqref{eq:rankLa} as a vector and subsequently accumulate these values
in \eqref{eq:rankL}. Our methods are implemented in two algorithms
for computing a new search direction.

\section{Algorithms}
\label{sec:algorithms}

This section describes algorithms for computing search directions
based on our estimates of the Jacobian matrix. The model weights are
updated according to $ \bkoex{w} = \bkex{w} + \bkex{s} $. Here the square of the diagonal matrix $ \bkex{d} = \text{diag}(\bkex{D}) $ is stored in a vector a squared accumulated gradients
\begin{equation}
    {\bkex{d}}^2 = \sum_{s=1}^k \b{g}^{(s)} \texttt{.*} \b{g}^{(s)}, 
\end{equation}
where $ \texttt{.*} $ represents element-wise multiplication. Similarly,
$ \texttt{./} $ will stand for element-wise division and $ \texttt{.}\sqrt{~} $
for the element-wise square root. 
Algorithm \ref{alg:alg1} implements the rank-1 estimate from Section
\ref{sec:rank1}.

\begin{algorithm}[tb]
   \caption{NLLS1 (Nonlinear Least Squares Rank-1)}
   \label{alg:alg1}
\begin{algorithmic}
   \STATE {\bfseries Initialize:} 
   $ k = 0 $, $ f^{(k)} = 0 $,
   $ \bkex{j} = \texttt{zeros}(n) $,
   $ \bkex{d} = 1\times 10^{-5} \texttt{ones}(n) $, $ \gamma = B $,
   $ \delta = \mathcal{O}( \sqrt{\frac{L}{4\gamma}} ) $,
   $ \alpha = 5\times10^{-3} $  
   \FOR{$\texttt{Epoch}=0,1,\ldots$ } 
    \FOR{$b=1,2,\ldots,B$ }  
    \STATE{ $f^{(k)} \gets f^{(k)} + \frac{\|\bkex{r}_b\|^2}{L} $ }
    \texttt{~~~~~~\#Loss~accum.}
    \STATE{ $\b{j}^{(k)} \gets \b{j}^{(k)} + \bkex{g}_b $} \texttt{~~~~~~~~~\#Jac.~accum.}
    
    \STATE{ $\b{d}^{(k)} \gets \b{d}^{(k)} + \bkex{g}_b \texttt{.*} \bkex{g}_b $ }
    \texttt{~~~\#Diag.~accum.}
    \STATE{ $\bkex{v} =  \frac{\delta}{\sqrt{f^{(k)}}} \bkex{j} $} 
    \STATE{\texttt{\#Representation:~H~and~H}$^{-1}$\texttt{(cf.~\eqref{eq:search})}}
    \STATE{\texttt{\#H=}$\frac{1}{\alpha}$
    \texttt{D+vv}$\tp$}
    \STATE{\texttt{\#H}$^{-1}$\texttt{=}$\alpha$\texttt{D}$^{-1}$\texttt{-}
    $\frac{1}{\alpha+\texttt{v}\tp\texttt{D}^{-1} \texttt{v}}\texttt{D}^{-1}\texttt{v}\texttt{v}\tp\texttt{D}^{-1}$}
    \STATE $ \bkex{s}_1 = -\alpha \bkex{g}_b\texttt{./}(\texttt{.}\sqrt{\bkex{d}})   $
    \STATE $ \alpha^{(k)}_1 = {\bkex{v}}\tp \bkex{s}_1 $
    \STATE $ \bkex{s}_2 = \alpha \bkex{v}\texttt{./}(\texttt{.}\sqrt{\bkex{d}})$
    \STATE $ \alpha^{(k)}_2 = {\bkex{v}}\tp \bkex{s}_2 $
    \STATE $ \bkex{s} = \bkex{s}_1 - \frac{\alpha^{(k)}_1}{1+\alpha^{(k)}_1} \bkex{s}_2 $
    \STATE $ \bkex{w} \gets \bkex{w} + \bkex{s} $
    \STATE $ k \gets k + 1 $
    \ENDFOR
   \ENDFOR
\end{algorithmic}
\end{algorithm}

Note that in Algorithm \ref{alg:alg1} part of the problem specifications,
i.e. $ B $ (number of batches) and $L$ (number of data pairs in a batch)
determine the order of magnitude for the parameter $\delta$. Otherwise,
we initialize $ \bkex{d} = 1\times 10^{-10} \texttt{ones}(n) $ and
$\alpha = 5 \times 10^{-2} $. Moreover, if Jacobian accumulation were
disabled, meaning that $ \bkex{j} = \b{0} $, then $ \bkex{s}_2 = \b{0}  $ and the search directions
$ \bkex{s} $ reduce to a basic version of the \emph{Adagrad} algorithm \cite{DuchiHazanSinger11}.
The method of Section \ref{sec:rankL} can be implemented similarly in
another algorithm. 
We will refer to the implementation of the rank-L
Jacobian approximation as {\small NLLSL}. 

Note that our algorithms can be implemented using element-wise
multiplications, divisions, and square roots only, and are thus applicable
to large problems. 

\section{Numerical Experiments}
\label{sec:experiments}

This section describes numerical experiments on three different optimization
problems using the least squares loss. We compare our implementations
of {\small NLLS1} (Algorithm \ref{alg:alg1}) and {\small NLLSL} with the widely 
used methods \emph{SGD}, \emph{Adam} \cite{KingmaBa15} and \emph{Adagrad} \cite{DuchiHazanSinger11}. The experiments are re-run 5 times for which
the average loss evolution is plotted. On the x-axis are the number of
epochs for a given experiment, which correspond to the number of
data passes over the full dataset (each epoch the dataset is composed
of $B$ batches with $L$ (feature, label) pairs). The algorithms are
implemented in Phyton 3.7 with TensorFlow 2.4.0 \cite{tensorflow} to compute the machine learning models and their derivatives. The experiments are carried out
on a MacBook Pro @2.6 GHz Intel Core i7 with 32 GB of memory. 
The codes are freely available at 
\begin{center}
 \texttt{\url{https://github.com/johannesbrust/SNLLS}}
\end{center}
    

\subsection{Experiment 1}
\label{sec:ex1}
\begin{figure}[ht]
\begin{center}
\centerline{\includegraphics[width=0.9\columnwidth]{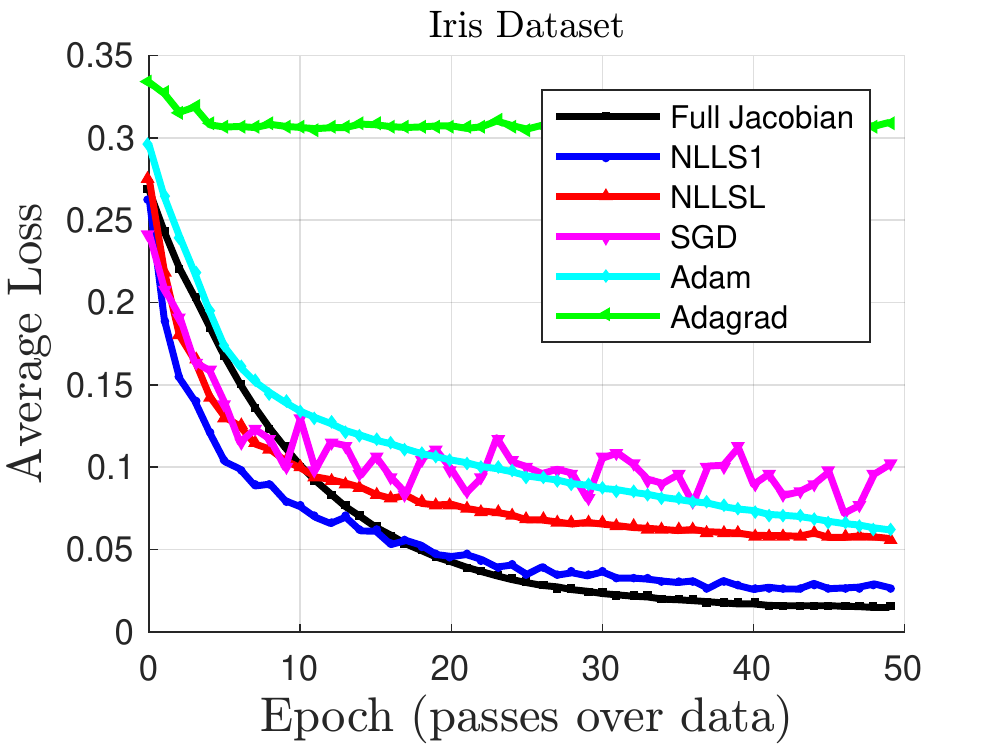}}
\caption{Comparison of 6 algorithms for the classification of 3 types of Iris
plants (dataset: \cite{DuaGraff19}). Observe that the ``Full Jacobian'' method
reaches the lowest average loss as expected, since this method explicitly
computes all Jacobian derivatives. However, the proposed rank-1 Jacobian
approximation (NLLS1) achieves similar low losses, without explicitly computing
the derivatives. The experiment outcomes are averaged over 5 runs.
}
\label{fig:ex1}
\end{center}
\vskip -0.4in
\end{figure}

This experiment compares 6 algorithms on a small classification
problem using the Iris flower dataset. A fully connected network with three dense
layers, \texttt{softmax} thresholding and \texttt{relu} activation is used.
The dataset contains 3 different flower classes, which are characterized
by 4 features. The number of optimization weights is $ n = 193 $,
with data sizes $ L = 96 $ and $ B = 4 $. Since the data is relatively
small, we additionally include the ``Full Jacobian'' method, which explicitly
computes the entire Jacobian in \eqref{eq:derivsJ}. Note however that unless $n$ and $L$ are
small, computing full Jacobian matrices is not practical. Adam obtained good results using default TensorFlow parameters, while 
using a learning of $1.0$ improved SGD. Adagard used the same learning rate
of $1.0$ as SGD. The parameter in Algorithm \ref{alg:alg1} is
$\delta = 0.8 \approx \frac{\sqrt{ L / 4 B }}{3} $, and the outcomes are in Figure \ref{fig:ex1}.

\subsection{Experiment 2}
\label{sec:ex2}
The movie recommendation dataset
contains 100,000 (movie title, rating) pairs. A fully connected model
with two dense layers and \texttt{relu} activation is used for rating
predictions. Movie titles and ratings are embedded in two preprocessing
layers of dimension 32. The total number of optimization weights 
is $n=116,641$. The data sizes are $ L = 8,192 $ and $B=13$.
Since this problem is relatively large, computing full Jacobian matrices
is not practical. Therefore the ``Full Jacobian" method is not applicable in
this experiment. However, all other algorithms are able to scale to the problem
size. We use the default TensorFlow parameters for all algorithms, expect Adagrad
which obtained better results with a learning rate of $0.1$. We set the
parameter $ \delta = 20 \approx 2 \sqrt{L/4B} $.

\begin{figure}[ht]
\begin{center}
\centerline{\includegraphics[width=0.9\columnwidth]{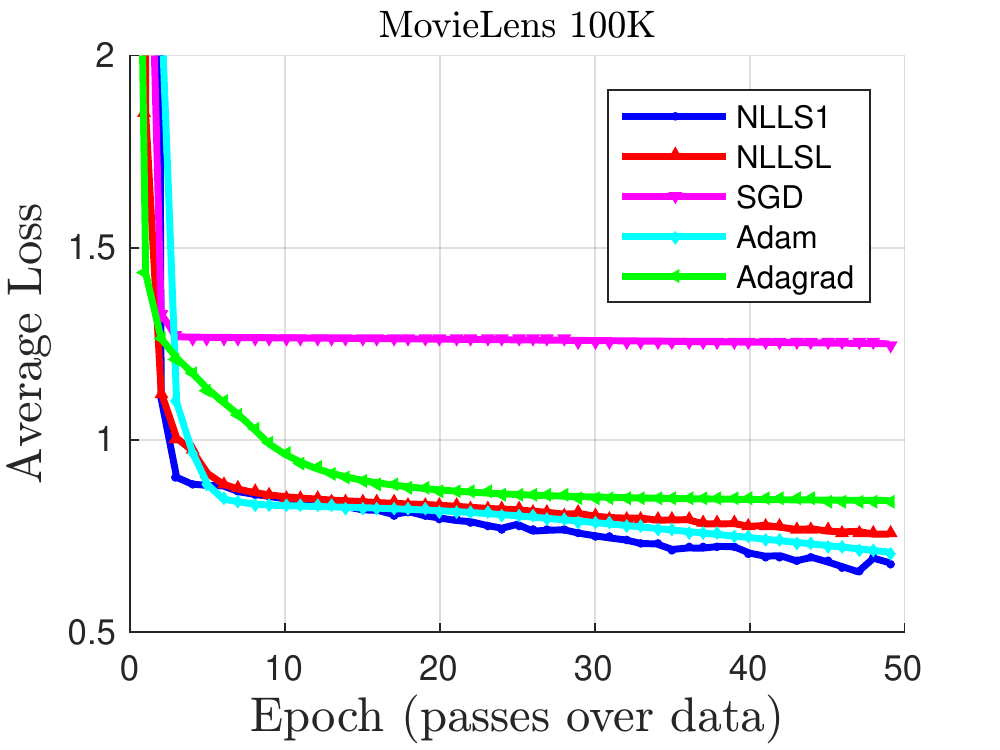}}
\caption{Comparison of 5 algorithms for training a movie recommender model on the MovieLens 100K dataset \cite{HarperKonstan15}. The proposed algorithm
{\small NLLS1} obtains the overall lowest losses. The outcomes are calculated
as the average of 5 runs.
}
\label{fig:ex2}
\end{center}
\vskip -0.4in
\end{figure}

\subsection{Experiment 3}
\label{sec:ex3}

In Experiment 3, the algorithms are applied to a large autoencoding
model. The Fashion MNIST dataset of 60,000 $ 28 \times 28 $ pixel images is used.
The model consists of a dense layers with \texttt{relu} activation for 
the decoding (with embedding dimension 64) and dense layers with 
\texttt{sigmoid} activation for the decoding step. The total number of
weights are $ n = 101,200 $ and the data consists of $ L = 25,088 $ and
$ B = 1,875 $. We use the default TensorFlow parameters for Adam and set the
learning rate to $50$ for SGD and Adagrad. We set
$ \delta = 0.9 \approx 0.5 \sqrt{L/4B} $.
\begin{figure}[H]
\begin{center}
\centerline{\includegraphics[width=0.9\columnwidth]{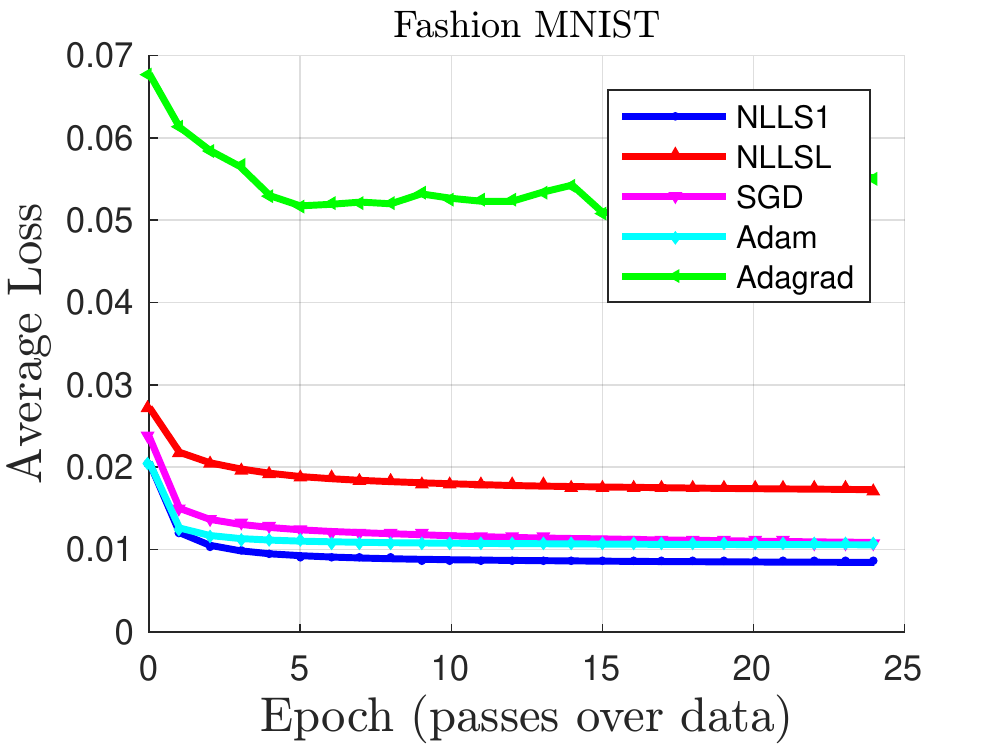}}
\caption{Comparison of 5 algorithms for training an autoencoder on the
Fashion MNIST dataset \cite{XiaoRasulVollgraf18}. {\small NLLS1} 
achieves the overall lowest loss values. The results are averaged
over 5 runs.
}
\label{fig:ex3}
\end{center}
\vskip -0.4in
\end{figure}

\section{Conclusions}
\label{sec:conclusions}

This article develops algorithms that approximate Jacobian information
in order to improve search directions in 
nonlinear least squares loss functions. By using the fact that the number
of model parameters typically exceeds the number of data pairs in
a batch of the dataset, we propose computationally effective
methods for computing search directions. In numerical experiments,
including large-scale applications, our proposed
algorithms perform well when compared to the state-of-the-art such as
Adam or Adagrad.

\bibliography{references}
\bibliographystyle{icml2021}





\end{document}